**Title:** A noise based novel strategy for faster SNN training

**Author names and affiliations:**

Chunming Jiang[a], Yilei Zhang[a]

a: department of mechanical engineering, University of Canterbury, New Zealand.

**Corresponding author:** Yilei Zhang    **Email address:** yilei.zhang@canterbury.ac.nz

**Abstract:** Spiking neural networks (SNNs) are receiving increasing attention due to their low power consumption and strong bio-plausibility. Optimization of SNNs is a challenging task. Two main methods, artificial neural network (ANN)-to-SNN conversion and spike-based backpropagation (BP), both have their advantages and limitations. For ANN-to-SNN conversion, it requires a long inference time to approximate the accuracy of ANN, thus diminishing the benefits of SNN. With spike-based BP, training high-precision SNNs typically consumes dozens of times more computational resources and time than their ANN counterparts. In this paper, we propose a novel SNN training approach that combines the benefits of the two methods. We first train a single-step SNN(T=1) by approximating the neural potential distribution with random noise, then convert the single-step SNN(T=1) to a multi-step SNN(T=N) losslessly. The introduction of Gaussian distributed noise leads to a significant gain in accuracy after conversion. The results show that our method considerably reduces the training and inference times of SNNs while maintaining their high accuracy. Compared to the previous two methods, ours can reduce training time by 65%-75% and achieves more than 100 times faster inference speed. We also argue that the neuron model augmented with noise makes it more bio-plausible.

**1. Introduction**

Spiking neural networks (SNNs) recently attracted increasing attention due to their biological plausibility. The SNN incorporates the concept of time into the model, and neurons in the SNN receive input spike trains that either increase or decrease their membrane potential. Through temporal accumulation, membrane potential may reach a specific firing threshold and neurons transmit information by firing discrete spike trains to neurons in the next layer. These characteristics emulate the information transmission and processing in the brain. It is therefore regarded as the next-generation neural network (Tavanaei et al., 2019).

Since SNNs use non-differentiable spikes as information carrying agents, gradient-based backpropagation (BP) that uses gradients to optimize synaptic connections and neuron parameters in ANNs is not directly applicable in SNNs. Thus, one of the main challenges is to train and optimize the network parameters in SNNs. At present, the available methods for training SNNs can be divided into three categories: (1) unsupervised learning, (2) indirect supervised learning, (3) direct supervised learning.

In the first approach, weights are modulated to mimic synaptic interactions between biological neurons. A classic example is the spike time-dependent plasticity (STDP) (Diehl & Cook, 2015; Kheradpisheh et al., 2016; Querlioz et al., 2013). However, due to the reliance on local neuronal activity rather than global supervision, STDP-based unsupervised algorithms have been limited to training shallow SNNs and can only produce low accuracy on complex datasets (Han et al., 2020; Masquelier & Thorpe, 2007; Srinivasan et al., 2018).

In the second approach, an ANN model is first trained and then converted to a SNN with the same network structure, where the firing rate of the SNN neuron is approximated as the analog output of the ANN neuron. The ANN-to-SNN conversion has produced state-of-the-art (SOTA) performance in image recognition tasks (Roy et al., 2019).

The last approach is direct supervised learning, which uses a similar gradient descent technique used in ANNs to train SNNs directly. SpikeProp (Bohte et al., 2002) was the first BP-based supervised learning method for SNNs that uses a linear approximation to overcome the SNNs' non-differentiable threshold-triggered firing mechanism. Further works include Tempotron (Gutig & Sompolinsky, 2006), ReSuMe (Ponulak & Kasiński, 2010), and SPAN (Mohemmed et al., 2012). However, they could only be used for training single-layer SNNs. The SuperSpike (Zenke & Ganguli, 2018) algorithm is designed to improve the accuracy of SNNs by optimizing the timing of the spikes. The algorithm uses a variant of backpropagation to adjust the weights of the network based on the timing of the spikes. A method proposed by (Huh & Sejnowski, 2018) also optimizes the spiking network dynamics for general supervised tasks on the time scale of individual spikes as well as the behavioral time scales. However, these methods still have a large performance gap with ANN when optimizing deep SNNs. A surrogate gradient algorithm proposed by (Wu et al., 2018) introduces a differentiable surrogate function to approximate the derivative of spiking activity. It executes spatio-temporal BP in the training phase and is widely applied to train deep SNNs.

Although the ANN-to-SNN conversion and surrogate gradient-based algorithm can train deep SNNs, there are some limitations. For the ANN-to-SNN conversion, training an ANN model is fast. Nevertheless, the approach requires considerable inference time (from hundreds to thousands of time steps) to approximate the analog outputs (Cao et al., 2015; Hunsberger & Eliasmith, 2015; Rueckauer et al., 2017; Sengupta et al., 2019; Stöckl & Maass, 2021), which leads to high memory consumption, larger latency and decreased energy efficiency, diminishing the benefits of SNNs (Deng & Gu, 2021; Roy et al., 2019; Severa et al., 2019). For the surrogate gradient-based direct training method, although it is possible to train SNNs with arbitrary time steps, the fewer the time steps, the lower the accuracy of the trained model would be. Training high accuracy SNNs with this approach often requires many times more training time and computational resources than training ANNs.

In this paper, we propose a novel SNN training method that combines the conversion concept and direct training using the surrogate gradient. The method consists of two phrases: single-step SNN(T=1) direct training and conversion to multi-step SNN(T=N). Specifically, during the training phase, SNN(T=1) augmented with Gaussian noise is trained by surrogate gradient-based BP and then converted losslessly to a multi-step SNN model to promote its generalization capability. Our training technique greatly reduces not only training and inference time but also achieves a high accuracy, which significantly improves the operating efficiency of SNN.

The following summarizes the primary contributions of this paper:

1) We propose a novel noise based SNN training algorithm, which speeds up the training and inference time of SNN.

2) We compare our method's training and inference time with those of current methods. The experiments demonstrate that our method is 3-5 times faster for training than the surrogate-gradient based method, and more than 100 times faster than the ANN-to-SNN conversion for inference.

3) We argue that introducing noise in SNN has biological plausibility.

## 2. Methods

### 2.1. Leaky Integrate-and-Fire model

The leaky Integrate-and-Fire (LIF) model is a fundamental unit in SNNs. It is a simplified representation of biological neurons that describes the non-linear relationship between input and output. The LIF neuron receives spikes over a specific period and it integrates them into its membrane potential, whose dynamics are governed by

$$H(t) = \lambda \cdot V(t-1) + \sum_i w_i \cdot S_i(t) \tag{1}$$

$$S(t) = \begin{cases} 1, & H(t) > V_{th} \\ 0, & H(t) \leq V_{th} \end{cases} \tag{2}$$

$$V(t) = H(t)\big(1 - S(t)\big) + V_{\text{reset}} \cdot S(t) \tag{3}$$

where $H(t)$ and $V(t)$ represent the membrane potentials before and after triggering a spike at time $t$, respectively. $\lambda$ represents the decay factor with a value of 0.5. $S(t)$ denotes the output of a neuron at time $t$, which equals 1 if there is a spike and 0 otherwise. $w_i \cdot S_i(t)$ is the weighted input of $i$-th neuron in the last layer at time step $t$. When the membrane potential of the LIF neuron reaches the firing threshold $V_{th}$ (= 1), the neuron fires one spike and the membrane potential is reset to the resting potential $V_{reset}$ (here is 0).

**2.2. Surrogate gradient method**

The method of surrogate gradient defines the derivative of the threshold-triggered firing mechanism using a surrogate derivative, which was proposed and developed (Neftci et al., 2019; Wu et al., 2018) to overcome the non-differential property of spiking neurons. It approximates the gradient of the spike function, which allows for the computation of gradients and the application of traditional optimization algorithm stochastic gradient descent (SGD). After the training process is completed, the final weights obtained from the surrogate gradient method can be used to produce spikes in the actual SNN, allowing it to function as intended. The surrogate gradient method has been shown to be effective in training SNNs on a variety of tasks and has been widely used in the development of SNN-based systems.

The form of surrogate gradient function varies. In this paper, we use Equation (4) ($a$=3) to approximate the gradient of spike function, which is plotted in Figure 1.

$$y(x) = \frac{a}{2\left(1+\left(\frac{\pi}{2}a(x-V_{th})\right)^2\right)} \quad (4)$$

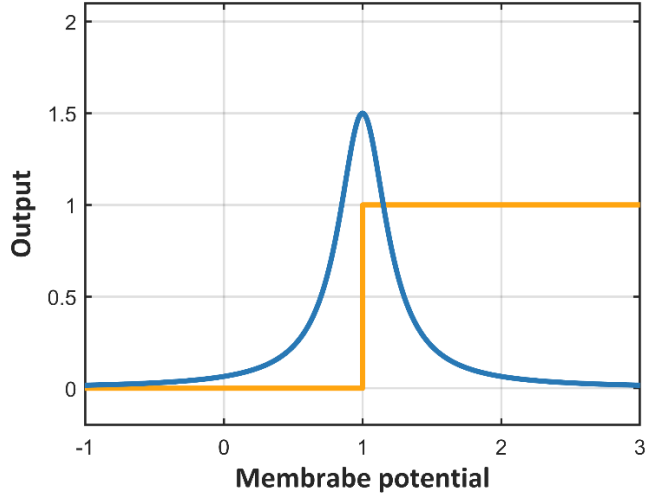

Fig.1 Surrogate gradients for the Heaviside function used in our experiments: The orange line represents the step function responsible for spike generation while the blue line approximates its derivative (surrogate gradient).

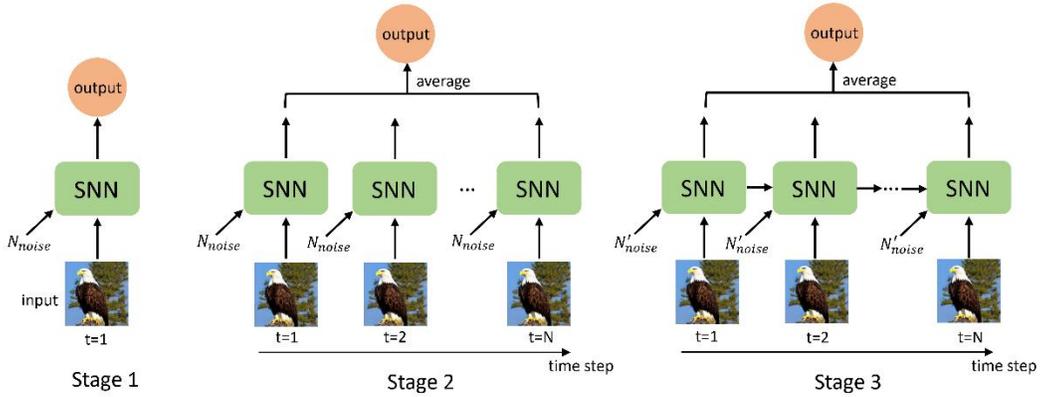

Fig.2 Three stages of our method to train a SNN model. Stage 1, single-step SNN(T=1) training with noise distribution $N_{noise}$. Stage 2, change the value of simulation time from T=1 to T=N. $N_{noise}$ varies over the time step t. Stage 3, establish the temporal correlation among N time steps.

### 2.3. Training single-step SNN(T=1) and converting to multi-step SNN(T=N)

We first train the single-step SNN(T=1) and then convert it to a multi-step SNN(T=N). When the total simulation step T=1, the time dimension disappears and the network propagates forward only once. Consequently, the single-step SNN is actually an ANN with a Heaviside step function used as the activation function. Equation (1) is formulated as

$$H = \sum_i w_i \cdot S_i \tag{5}$$

Comparing Equations (1) and (2), the output of the multi-step LIF neuron depends on both the input and the accumulated potential, while the output of a single-step LIF neuron depends only on the input.

Due to the absence of the potential accumulation term in Equation (5) compared with Equation (1), we introduce a noise distribution $N_{noise}$ representing the missing accumulated membrane potential during training the single-step SNN in order to perform the conversion into a multi-step SNN later, as Equation (6) shows. In particular, we assume that $N_{noise}$ is a Gaussian distribution and distributes in each layer of the network independently. Thus, the dynamic of the neuron in a single-step SNN could be described by

$$H = N_{noise} + \sum_i w_i \cdot S_i \tag{6}$$

Our method consists of three following stages:

**Stage 1:** Train the single-step SNN(T=1) with Gaussian noise (see stage 1 in Figure 1) by the surrogate gradient. Notably, the added noise $N_{noise}$ varies randomly for each training iteration. It is not the same frozen noise over iterations. After training is completed, all parameters except $N_{noise}$ are frozen and there is no gradient calculation and any training in stages 2 and 3.

**Stage 2:** Change the value of simulation time from T=1 to T=N by simply modifying the value of T. When T=N, the input is fed N times in the time dimension into the SNN model, and the noise at each time step $t$ is randomly varying. We average the outputs of N times as the final output of SNN model. (see stage 2 in Figure 1).

**Stage 3:** Add the potential accumulation term using Equation (7) to establish the temporal correlation among different time steps (see stage 3 in Figure 1). The dynamic of SNN model after step 2 is formally different from the real SNN's dynamic because it lacks the process of potential accumulation and the temporal correlation among different time steps. Consequently, we must add the item $\lambda \cdot V(t-1)$ to keep the formal consistency with Equation (1). We decompose $N_{noise}$ into:

$$N_{noise} = N(\lambda \cdot V(t-1)) + N'_{noise} \tag{7}$$

$N_{noise}$ could be represented as the addition of two Gaussian distribution items: the accumulated membrane potential distribution and the new noise distribution.

For the first item, we normalize $\lambda \cdot V(t-1)$ according to Equations (8)-(10) to approximate a Gaussian distribution $N(\lambda \cdot V(t-1))$. The reason is that we have added the Gaussian noise distribution $N_{noise}$ in stage 1 and assume that it has an approximate distribution with the potential accumulation term. But the actual potential accumulation term $\lambda \cdot V(t-1)$ does not necessarily obey a Gaussian distribution, so we constrain the potential accumulation term in stage 3 to an approximate Gaussian distribution by the item $N(\lambda \cdot V(t-1))$ in Equation (7). In this way, we can introduce the noise term $N_{noise}$ for training with T=1 in stage 1 and replace it with the potential accumulation term after training is completed.

$$A = \lambda \cdot V(t-1) \tag{8}$$

$$\hat{A} = \frac{A - \mu}{\sigma} \tag{9}$$

$$N(\lambda \cdot V(t-1)) = \frac{\hat{A}}{\alpha \cdot max(abs(\hat{A}))} + \beta \tag{10}$$

Equation (9) converts the membrane potential distribution to a standard normal distribution approximately. μ and σ are the mean and standard deviation of $A$, respectively. Equation (10) guarantees that the potential distribution in each layer have the mean $\beta$ and distributes between the interval $(-1/\alpha, 1/\alpha)$. By changing the values of $\alpha$ and $\beta$, we are able to change the mean and range of the distribution. ($max()$ and $abs()$ represent taking the maximum value and absolute value, respectively.)

For the second item $N'_{noise}$, we also keep it as a random Gaussian distribution. The mean and range of $N'_{noise}$ also depends on $\alpha$ and $\beta$, because we need to ganrantee that the addition of two items in Equation (7) almost have the same mean and range as $N_{noise}$ to avoid conversion loss. The reason why we keep the random noise distribution $N'_{noise}$ is that introducing the noise plays the important role in promoting accuracy, which have been discussed in detail in section 3.

It is well known that a SNN model with more simulation steps T can increase performance. However, training a SNN with large T directly would increase not only the training and inference time but also the memory by T folds, so it is not very practical. In our proposed method, the whole training process is completed in stage 1. In stages 2 and 3, things we need to do are to change the value of simulation time from T=1 to T=N and to replace $N_{noise}$ using Equation (7). The approach allows training SNNs using only one time step, which make it possible to train models with much less memory cost.

## 3. Results and discussions

We conduct our experiments as Table 1 shows. SNNs are trained with MSE loss and Adam (Kingma & Ba, 2014) optimizer. The initial learning rate is set to 1e-4. The cosine annealing warm restart (Loshchilov & Hutter, 2016) learning rate scheduler with $Tmax = 100$ adjusts the learning rate over training. Unless specified, all results are generated by default for $\alpha = 4$, $\beta = 0.5$, and $N_{noise}$ in the range [0, 1].

Table 1 Network structures and training epoch for different datasets.

| Dataset | Epoch | Network structure |
|---------|-------|-------------------|
| MNIST | 100 | 64C3-AP2-128C3-AP2-128C3-AP2-512FC-10FC |
| Fashion -MNIST | 200 | 64C3-AP2-128C3-AP2-128C3-AP2-512FC-10FC |
| CIFAR-10 | 1000 | VGG-16 |
| CIFAR-10 | 800 | ResNet-18 |

Note: nC3—Convolutional layer with n output channels, kernel size = 3 and stride = 1, AP2—2D average-pooling layer with kernel size = 2 and stride = 2, FC—Fully connected layer.

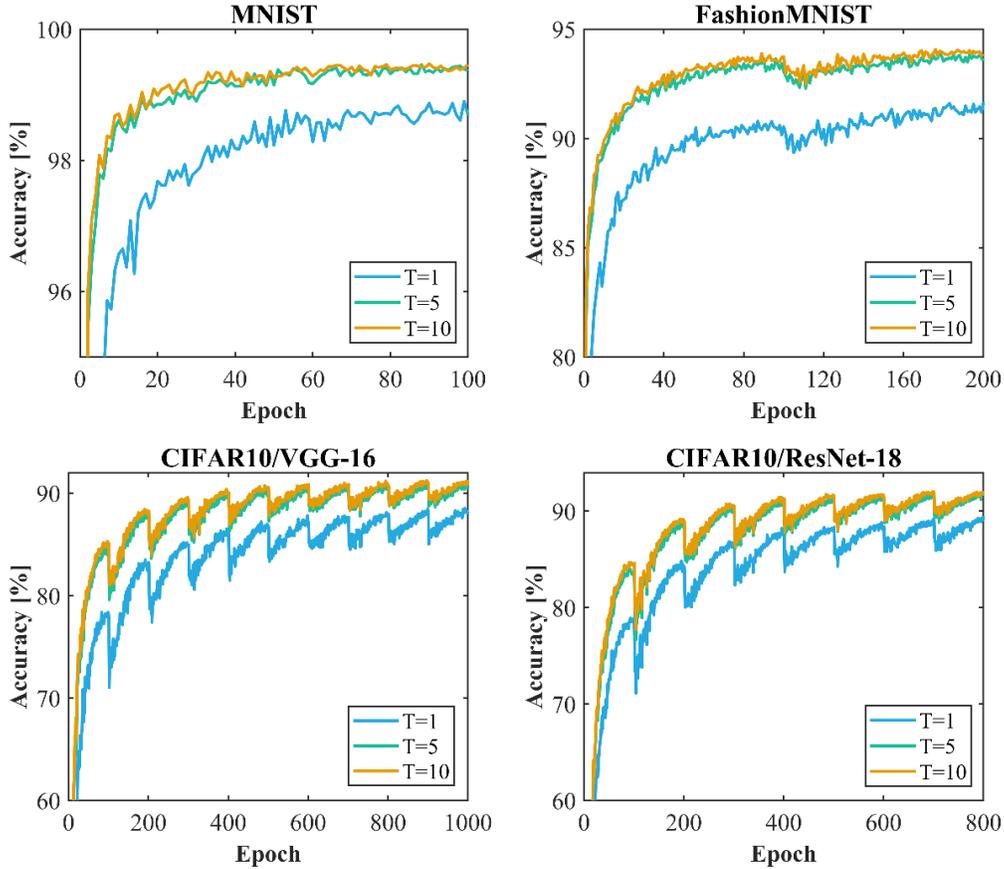

Fig.3 Inference accuracy of models on different datasets with T = 1, 5, 10 while training with $N_{noise}$.

### 3.1. Inference accuracy

In Figure 3, we plot the inference accuracy of single-step SNNs(T=1) on different datasets and the inference accuracy of multi-step SNNs with total simulation step T=5 and T=10, respectively. It can be shown that as T increases, the accuracy of the SNN improves dramatically. The simulation step T could be directly changed to any values in real time. The fluctuation of curves in the figure is because the scheduler with warm restart resets the learning rate every 100 epochs.

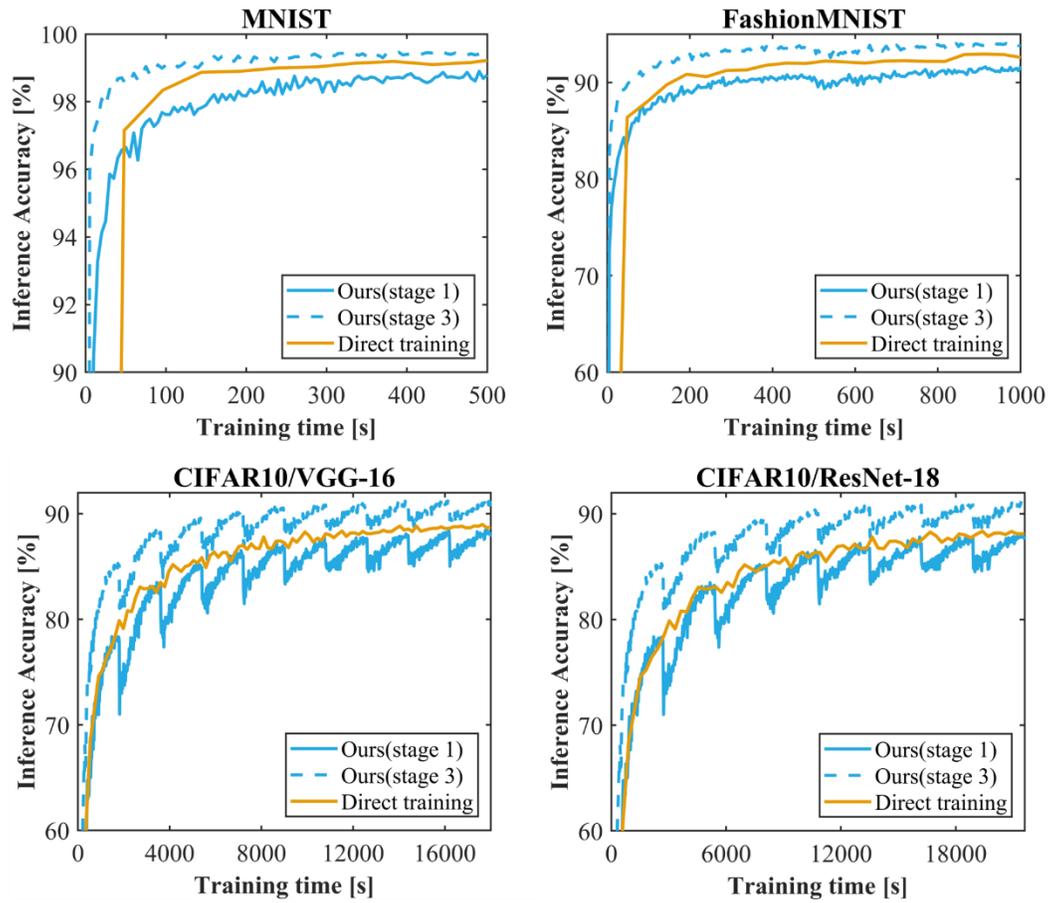

Fig.4 Comparison of training time for training SNN(T=10) with direct training method and our method. Ours (stage 1) is the inference accuracy after step 1; Ours (stage 3) is the inference accuracy after stage 3.

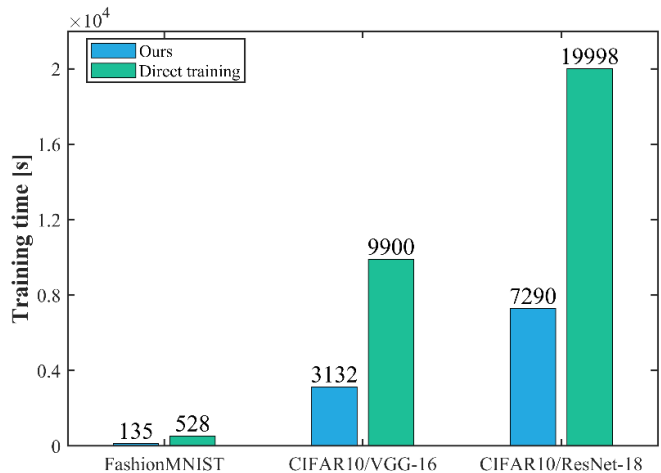

Fig.5 Comparison of training time for training SNN(T=10) by direct training method versus our method. The benchmark inference accuracy of the three models is 92%, 88%, and 90%, respectively.

### 3.2. Comparison of training and inference time with related work

In Figure 4, we compare the training speed of the direct training method to train SNN(T=10) models with our method training. The X-axis represents the training time and Y-axis represents the inference accuracy. The direct training method directly trains SNN models with T=10 by the surrogate gradient, while our method executes stage 1-3 described in section 2.3. Clearly, our method is substantially faster than directly training method with the same training time. To intuitively assess the difference in training time, we selected a benchmark inference accuracy for each model and halted training when the benchmark inference accuracy was achieved. The benchmark inference accuracy of the three models is 92%, 88%, and 90%, respectively. As shown in Figure 5, on FashionMNIST our method takes only 135s to reach the benchmark accuracy when direct training takes 528s, which saves about 75% of the time. In CIFAR10/VGG-16, our method requires 3132s, whereas the other requires 9900s, a time savings of about 70%. In CIFAR10/ResNet-18, our model requires 7290s while direct training requires 19998s, a time savings of approximately 65%.

We consider that the addition of noise during training makes the LIF neurons sensitive to the initial potential values inside the neurons at each time t, so that when we change the total simulation time T=1 to T=N in stage 2, the SNN model can be seen as the parallelization of N different models in the time dimension to get the accuracy improvement, which is similar to the concept of ensemble learning, except that the ensemble learning is the combination of several different models in space, while our method uses the ensemble of the same SNN models with different initial potential states in different time steps.

Therefore, our method only needs to train a SNN(T=1) model, which can be quickly converted into a model with SNN(T=10). It is obviously faster than directly training a SNN(T=10) model by the surrogate gradient. Also, since only single-step SNN(T=1) needs to be trained in our method, we can choose a larger batch size and thus achieve faster parallel training. It is more convenient and feasible for groups that lack sufficient computational resources.

For inference time, we compare current advanced methods listed in Table 2 with our method. As demonstrated, the accuracy of extending to multi-step SNN (no more than 10 time steps) is able to attain an approximate accuracy of ANN-to-SNN conversion methods. In contrast, ANN-to-SNN conversion requires hundreds to thousands of time steps, which is hundreds of times slower than our method. Compared with spike-based BP methods, our method also requires fewer time steps to reach close accuracy.

Table 2 Inference time comparison between our work and related work

| Author | Method | Inference time | MNIST | FashionMNIST | CIFAR10 |
|---|---|---|---|---|---|
| (Wu et al., 2019) | Spike-based BP | 12 | - | - | 90.53% |
| (Rathi & Roy, 2021) | Spike-based BP | 10 | - | - | 93.44% |
| (Cheng et al.) | Spike-based BP | 20 | 99.50% | 92.07% | 93.5% |
| (Severa et al., 2019) | Spike-based BP | 1 | 99.53% | - | 84.67% |
| (Sengupta et al., 2019) | ANN-SNN | 2500 | - | - | 91.46% |
| (Han et al., 2020) | ANN-SNN | 2048 | - | - | 91.36% |
| (Diehl et al., 2015) | ANN-SNN | 2500 | - | - | 91.89% |
| (Lee et al., 2020) | ANN-SNN | 50/100 | 99.59% | - | 90.95% |
| (Stöckl & Maass, 2021) | Hybrid | 500 | - | - | 92.42% |
| (Rathi et al., 2020) | Hybrid | 200 | - | - | 92.02% |
| Ours | Hybrid | 5 | 99.61%% | 93.89% | 91.82% |

| | Ours | Hybrid | 10 | 99.64% | 94.07% | 92.07% |

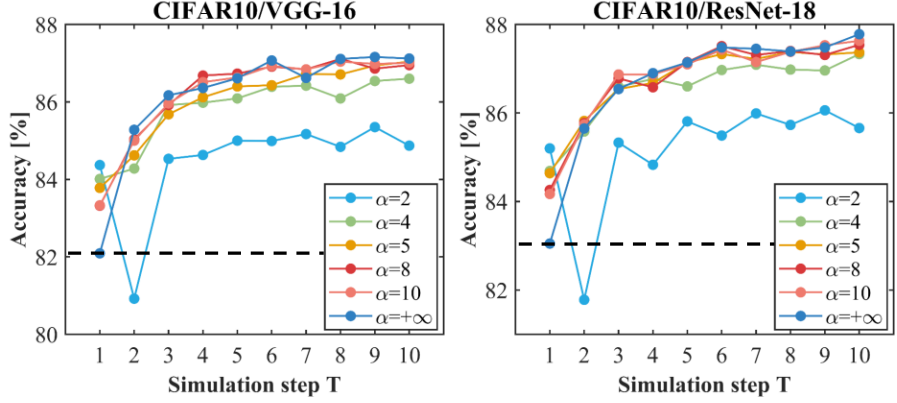

Fig.6 The impact of $\alpha$ on CIFAR10/VGG-16 and CIFAR10/ResNet-18. The black dotted line represents the accuracy of trained single-step SNN(T=1).

### 3.3. The impact of noise

With Equations (7) and (10), we know that the parameter $\alpha$ controls the range of noise fluctuation $N'_{noise}$. Here, we attempt to alter the value of $\alpha$ to observe how the model's performance varies.

We plot the accuracy in stage 3 for different values of $\alpha$ in Figure 6. When $\alpha$ is equal to 2, the potential range is [0,1], so the noise term $N'_{noise}$ in Equation (7) does not exist anymore. $N_{noise} = N(\lambda \cdot V(t-1))$ and the SNN does not include noise in stage 3. It can be seen that there is only a slight improvement in accuracy with increasing time steps. In contrary, when $\alpha > 2$, $N'_{noise}$ is present; the accuracy improvement is obvious and different $\alpha$ values make models converge to close accuracy. These results indicate that noise plays the vital role in enhancing the accuracy of conversion.

When $\alpha$ is positive infinity, membrane potential item disappears so $N'_{noise} = N_{noise}$; the curve represents the accuracy of stage 2. Comparing this curve with others, we can see that the conversion from stage 2 to stage 3 has minor accuracy gap according to the figure.

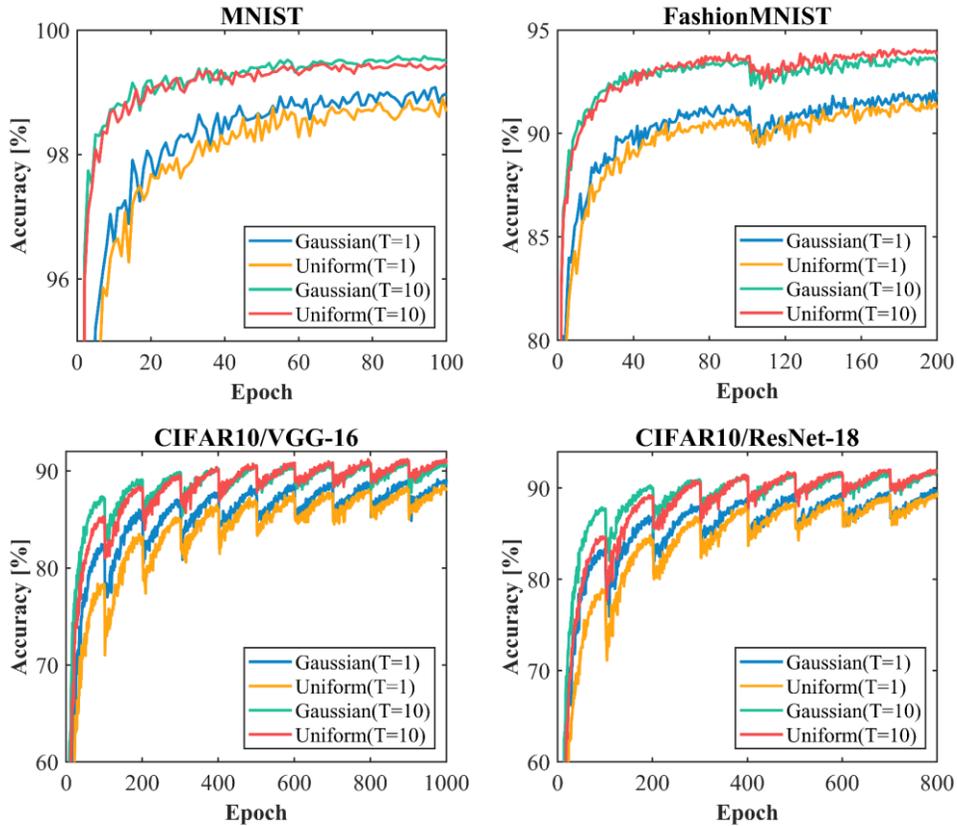

Fig.7 Inference accuracy of models on different datasets with T = 1, 10 while training using Gaussian noise and uniform noise, respectively.

### 3.4. The impact of noise type

In the previous sections, all our experiments were performed by training the SNN model with Gaussian noise. In order to investigate whether it is only the uniform noise that brings the improvement of model accuracy, we replace the Gaussian noise with uniform noise during training to observe the effect of the noise type on models. As shown in Figure 7, we trained CIFAR-10 with uniform noise on both VGG-16 and ResNet-18 models. Models trained by uniform noise behave the same as with uniform noise, i.e., the accuracy is significantly improved when models are extended to multi-step SNNs. For single-step SNNs, models with Gaussian noise reach the higher accuracy than those with uniform noise. But after conversion, the gap is not obvious anymore.

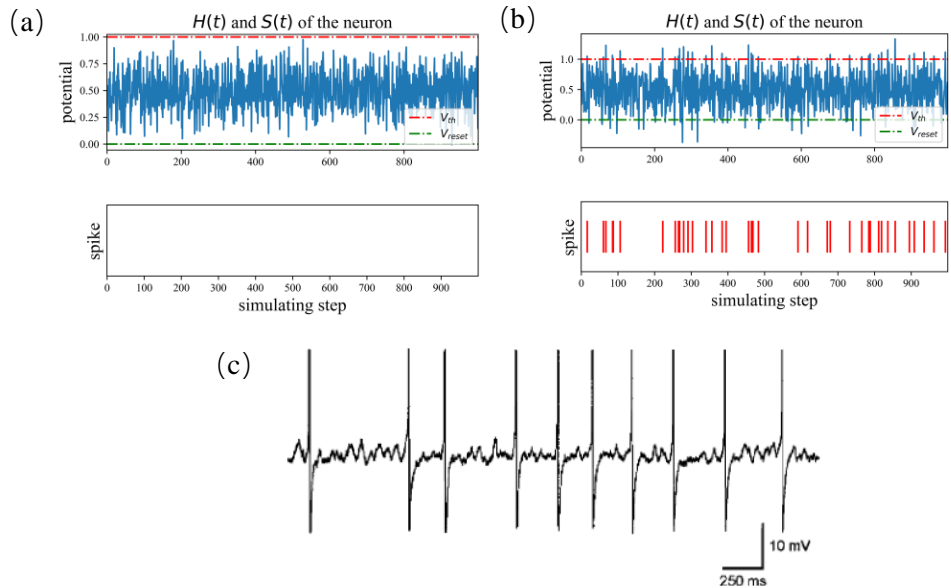

Fig.8 (a) Neural potential dynamic in the absence of input. (b) Neural potential dynamic when receiving input. (c) Subthreshold membrane potential oscillation. Source: Figure (c) is cited from (Boehmer et al., 2000).

### 3.5. Biological plausibility of uniform noise distribution in neuron

It is believed to be more biologically plausible when we keep some part of noise during conversion, as there exists lots of kinds of noise in biological neural system. We plot the dynamic of the spiking neuron in Figures 8(a) and (b). Figure 8(a) depicts the neural potential dynamic in the absence of input. The potential oscillates between the reset potential and the threshold, but no spikes fire. In Figure 8(b), when a neuron receives inputs, it begins to accumulate potential and fires spike. Such behavior is thought to be similar to the form of subthreshold neural oscillation mechanism in biological neurons. Neural oscillations are rhythmic patterns of activity generated by the neurological system (Başar, 2013). Many cognitive activities, including information transfer, perception, and memory, are believed to be related with neural oscillations. These oscillations are mostly caused by the interaction between individual neurons. Neural oscillation can emerge as oscillating membrane potentials or rhythmic action potentials in individual neurons. Subthreshold membrane potential oscillations are membrane oscillations that are below the firing threshold and hence cannot directly initiate action potentials. However, they can aid in sensory signal processing. As a result of subthreshold membrane potential oscillations, sensory systems, particularly for vision and smell, evolve. Subthreshold membrane potential oscillation (see Figure 8(c)) in the visual system helps process visual input and adjust to sensory input (Purves et al., 2008). Additionally, oscillatory activity influences excitatory postsynaptic potentials, refining post-neural activities (Desmaisons et al., 1999).

**4. Discussion and conclusion**

We have used the proposed method to handle static vision problems in previous sections. For dynamic vison problems, such as videos' data, which include time series information, we should feed all the information into the network at once and use a 3D convolutional network rather than a 2D convolution network to deal with the input.

In this paper, we propose a novel way of training SNNs that achieves accuracy improvement in multi-step SNNs by fitting the neural network to noise, which greatly spares the training and inference time of SNNs and allows fast training of SNNs with arbitrary simulation time compared to previous methods. Our approach combines the advantages of both direct training of SNN and ANN-to-SNN conversion. With a good balance of accuracy and training time, and a great saving of computational resources, this method can be used to train large SNNs quickly or SNN pre-training. The inclusion of noise is also proved to be more consistent with the dynamic mechanism of biological neurons. These points make our method promising for training deep SNN. In addition, we focus on computer vision classification tasks and convolutional neural networks in this paper. In subsequent work, it could be further validated our algorithms in more tasks, such as Natural Language Processing (NLP), and by combining SNNs with more structures, like Transformer.

**Acknowledgments**

We gratefully acknowledge the financial support provided by the University of Canterbury, New Zealand.